# A Neural Network Model for Construction Projects Site Overhead Cost Estimating in Egypt

Ismaail ElSawy[1], Hossam Hosny[2] and Mohammed Abdel Razek[3]

[1] Civil Engineering Department, Thebes Higher Institute of Engineering
Corniche Maadi, Cairo, Egypt

[2] Construction Engineering Department, Zagazig University
Zagazig, Egypt

[3] Construction and Building Department, Arab Academy for Science, Technology and Maritime Transport
Cairo, Egypt

**Abstract**
Estimating of the overhead costs of building construction projects is an important task in the management of these projects. The quality of construction management depends heavily on their accurate cost estimation. Construction costs prediction is a very difficult and sophisticated task especially when using manual calculation methods. This paper uses Artificial Neural Network (ANN) approach to develop a parametric cost-estimating model for site overhead cost in Egypt. Fifty-two actual real-life cases of building projects constructed in Egypt during the seven year period 2002-2009 were used as training materials. The neural network architecture is presented for the estimation of the site overhead costs as a percentage from the total project price.

## 1. Introduction

Applications of ANN (Artificial Neural Network) in construction management in general go back to the early 1980's. These applications cover a very wide area of construction issues. Neural network models have been developed internationally to assist the managers or contractors in many crucial construction decisions. Some of these models were designed for cost estimation, decision making, predicting the percentage of mark up, predicting production rate …etc.

The objective of this research is to develop a neural network (NN) model to assess the percentage of site overhead costs for building projects in Egypt. This can assist the decision makers during the tender analysis process.

Cost Estimating is one of the most significant aspects for proper functioning of any construction company. It is the lifeblood of the firm and can be defined as the determination of quantity and the prediction or forecasting, within a defined scope, of the costs required to construct and equip a facility.

The significance of construction cost estimating is highlighted by the fact that each individual entity or party involved in the construction process have to make momentous financial contribution that largely affects the accuracy of a relevant estimate. The importance and influence of cost estimating is supported by scores of researches.

Carty (1995) and Winslow (1980), for example, have documented the importance of cost estimating, mentioning it as a key function for acquiring new contracts at right price and hence providing gateway for long survival in the business. According to Larry, D. (2002) cost estimating is of paramount importance to the success of a project **[1]**.

Alcabes (1988), articulated that, estimating departments is responsible for the preparation of all estimates, estimating procedures, pricing information, check lists and applicable computerized programs. He also insists on the fact that accurate cost categorization, cost reporting, and profit calculation are the heart of the construction business. In order to achieve a financial engineered estimating methodology, it is imperative that different techniques should be evaluated **[3]**.

Hegazy and Moselhi (1995), conducted several surveys studies in Canada and the United States to determine the elements of costs estimation. The survey was carried out with the participation of 78 Canadian and U.S.A building construction contractors in order to elicit current practices with respect to the cost elements used to compile a bid proposal and to identify the types of methods used for estimating these elements. Their results indicated that direct cost and project overhead costs are estimated by contractors primarily in a detailed manner, which is contrary to the estimation of the general overhead costs and the markup [9].

Assaf, S. A. et al. (2001), investigated the overhead cost practices of construction companies in Saudi Arabia. They show how the unstable construction market makes it difficult for construction companies to decide on the optimum level of overhead costs that enables them to win and efficiently administer large projects [4].

Cost estimating models and techniques provides a well defined engineered calculation methods for the evaluation and assessment of all items of office overhead, project overhead, profit anticipation, total project cost estimation, and the assessment of overhead costs for construction projects that leads to competitive bidding in the construction industry [11].

This paper presents the steps followed to develop a proposed model for site overhead cost estimating. The necessary information and the required projects data were collected on two successive yet dependent stages:

I. Comparison between the list of site overhead factors collected from previous studies and the applied Egyptian site overhead list of factors that is adapted by the first and second categories of construction firms in Egypt; and
II. Collection of all required site overhead cost data for a sample of projects in Egypt to be used during the analysis phase and site overhead cost assessment model development.

## 2. Research Methodology

The findings from the survey conducted on all the previous researches served as key source in the identification of the main factors affecting site overhead costs for building construction projects. Based on an extensive review for the previous studies conducted in this area of work, the survey for such factors mainly include projects need for specialty contractors, percentage of sub-contracted works, consultancy and supervision, contract type, firm's need for work, type of owner/client, site preparation, projects scheduled time, need for special construction equipment, delay in projects duration, firms previous experience with projects type, legal environmental and public policies for the home country, projects cash-flow plan, project size, and projects location. Hence, the study shed a great deal of light on the area of site overhead costs for building construction projects in Egypt. Through seeking the experts opinions regarding the development of a list for the main factors affecting the building projects overhead costs. They will be used during the development of the model. Such factors were mainly identified based on the expert's opinions from selected groups of prominent industrial professionals and qualified academicians from the most prominent universities in Egypt. The principal objective of this survey study was to reinforce the potential model, based on the expert's opinions from the aforementioned expert professionals [12].

Expert opinion included the reviews from nineteen prominent industrial professionals and sixteen qualified academicians from the American University in Cairo and the Arab Academy for Science and Technology and Maritime Transport. Reviews from experienced industrial professionals were essential for developing the overall model as these professionals are directly associated with the leading Egyptian building construction firms.

Each expert from both contractor and academic background were approached based on their personnel experiences. Half of the responses were obtained via personnel interviews and the other half were obtained through delivering the questionnaire and collecting back the same, E-mail or Fax.

As this phase of seeking expert's opinion consist of the walk-through observations of the selected specified industrial professionals and academicians connected to the construction industry. These reviews provided us with qualified remarks and suggestions, which will lead to making the necessary alterations on the list of the previously identified overhead cost factors to make it adaptable to the Egyptian building construction industry market. This is an essential step to have a more firm and yardstick final model for the assessment of overhead costs for building construction projects, in Egypt [12].

## 3. Data Collection

This phase is divided into two stages; first stage is to perform a comparison between the overhead cost factors from the comprehensive literature study and the Egyptian construction industry. Hence, the main factors affecting site overhead costs can be clearly identified. The second stage is to collect data for 50 projects from several construction companies that represent the first and the second categories of construction companies, in Egypt [12].

3.1 The questionnaire

In the first section of the data collection process, a questionnaire is prepared to investigate the main factors affecting site overhead cost for building construction projects in Egypt.

The questionnaire consisted of three sections, the first section contained nine yes or no questions to confirm or eliminate any of the constituent factors that have been collected previously from the literature study. The second section is where the experts illustrate the factors currently accounted for by construction firms in Egypt. The third section is where the experts are asked for their own opinions for the factors that are not accounted for and should be considered in order to stroll with the construction industry in Egypt. The characteristics of the participating experts, the contractors and the academicians are setting the basis for the findings of this study. The mentioned characteristics of contractors include their personnel professional experience and size of the firm they are associated with. The distinctiveness of academicians described includes their designation, area of specialization and essentially their experience.

Experts for this extensive research are very scrupulously identified to obtain comprehensive and precise results. The highly capable experts were selected among the practicing, experienced contractor's professionals in Egypt and the highly qualified academicians from the two renowned universities not only in Egypt but in the entire region [12].

### 3.1.1 Academicians

Academicians are the professionals, who have strong influence on national research and scientific work. As part of this thesis, expert appraisals from faculty members belonging to Construction Engineering and Management or Civil Engineering fields from two prestigious universities in Egypt. The Academicians engaged for this research are icons from academia. Their expertises are articulated by the fact that, seventy percent of the respondents are either Professor or Associate Professor in the two renowned universities. Along with the aforementioned colossal qualification levels, the traits of the participating academic professionals include their experience, classified based on the number of years in academia. Thirty one percent of the interviewed experts are dedicating their services to the academic discipline from more than 20 years. Another forty four percent of the academic experts have 10-20 years of practicing experience (twenty five percent have from 15-20 years and nineteen percent have from 10-15) and twenty five percent have less than 10 years of professional experience in academia (Fig. 1) [12].

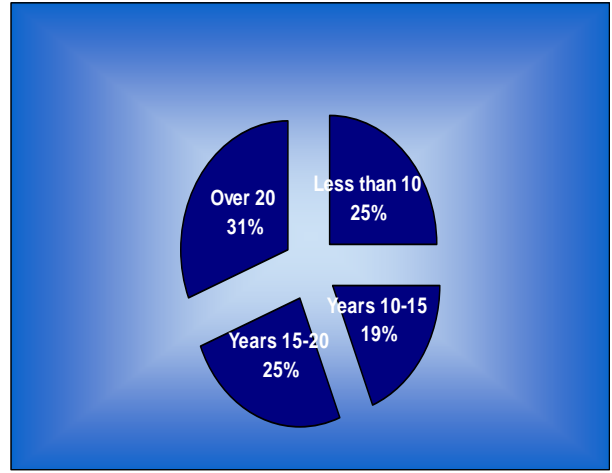

Fig. 1. Academicians Years of Experience. [12]

### 3.1.2 Contractors

The participating contractors (Cost Estimating Engineers) are highly experienced professionals from the construction industry. About fifty percent of the experts have more than 20 years of professional experience in the construction business. The remaining has experience less than 20 years. These vastly experienced industry professionals occupy senior and highly ranked administrative positions within their firms. Seventy percent of the experts are ranked as General Manager Engineers. The remaining thirty percent work as project cost estimation engineers. The participants work for successful construction firms belonging to the first and second categories. Twelve experts work for first category construction companies, five experts work for second category construction companies, and two experts work for a major construction consultancy firm all within Egypt, (Fig. 2).

The views of the contracting experts from firms of different grades were sought to get a more diversified & comprehensive review [12].

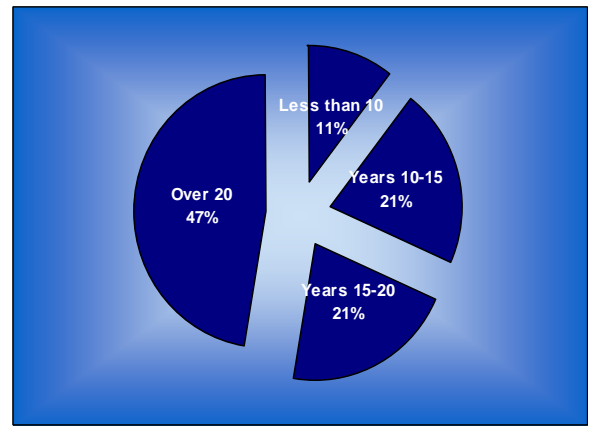

Fig. 2. Contractors Years of Experience. [12]

The analysis of the collected questionnaires illustrated that there is a difference between the factors that govern the assessment of building construction site overhead cost in Egypt and the international building construction industry trend. Many factors are not accounted for in Egypt due to its insignificance in the local market while it is a great contributor in both Europe and North/South America construction markets. Moreover, in Egypt there is a trend between contractors to combine two or more contributing items in one main factor. The academicians contravened with that behavior and characterized it to be an unprofessional attitude because it depends entirely on the person that is performing the task and his/her experience with the projects on hand (personalization). So after cross-matching and making the necessary alterations on the questionnaires collected from both the contractors and academicians in Egypt, a final list of factors were generated that represent both the parties and it can accurately represent the factors that contribute to building construction site overhead cost in the Egyptian construction market (Table 1) **[12]**.

Table 1: Factors Contributing to Construction Site Overhead Cost Percentage in Egypt

| | Factor |
|---|---|
| 1 | Construction Firm Category. |
| 2 | Project Size. |
| 3 | Project Duration. |
| 4 | Project Type. |
| 5 | Project Location. |
| 6 | Type-Nature of Client. |
| 7 | Type of Contract. |
| 8 | Contractor-Joint Venture. |
| 9 | Special Site Preparation Requirements. |
| 10 | Project need for Extra-man Power. |

## 4. Site Overhead Cost Data

A comparative analysis was performed between building construction site overhead cost and each constituent of site overhead regarding building construction projects, with the aid of (**52**) completed building construction projects. These projects were executed during the seven year period from 2002 to 2009. The comparison is made in terms of cost influence for each factor of projects site overhead on the percentage of projects site overhead cost in order to recognize and understand the governing relationship between each factor and the percentage of site overhead cost **[12]**.

It must be illustrated that for all the collected projects the adapted construction technology was typical traditional reinforced concrete technology. This may be due to the participating experts opinion, because that technology represents over (**95**%) of the adopted building construction technology in Egypt. Contrarily, if any specific construction technique is required for a certain project it must be accounted for by the construction firm cost estimating department in an exceptional manner **[12]**.

## 5. Comparative Analysis Results

The major and minor findings of the entire research were summarized in this part of the research. Based on the findings the current and further recommendations are developed as the base for further research in the very context of building construction projects overhead cost for the first and the second categories of construction companies, in Egypt **[12]**.

The analysis illustrated many facts that needed to be clarified and understood about the percentage of site overhead costs for building construction projects in Egypt. These facts will be the structure (backbone) for the development of a model for the assessment of site overhead cost as a percentage from the total contract amount for building construction projects, in Egypt. This can be simply summarized in the following two facts: **[12]**

A. Through the literature review and the expert's opinions potential factors that are found to influence the percentage of site overhead costs for building construction projects in Egypt, ten factors were identified.

B. The analysis of the collected data gathered from fifty-two real life building construction projects from Egypt during the seven year period from 2002 to 2009, illustrated that project's duration, total contract value, projects type, special site preparation needs and projects location are identified as the top five factors that affect the percentage of site overhead costs for building construction projects in Egypt.

## 6. Neural Network Model

The guidelines of N-Connection Professional Software version 2.0 (1997), users manual were used to obtain the best model. Moreover, for verifying this work the traditional trial and error process was performed to obtain the best model architecture **[11]**.

The following sections present the steps performed to design the artificial neural network model, ANN-Model.

Neural network models are generally developed through the following basic five steps **[8]**:

1. Define the problem, decide what information to use and what network will do;
2. Decide how to gather the information and represent it;
3. Define the network, select network inputs and specify the outputs;
4. Structure the network;
5. Train the network; and

6. Test the trained network. This involves presenting new inputs to the network and comparing the network's results with the real life results, (Fig. 3).

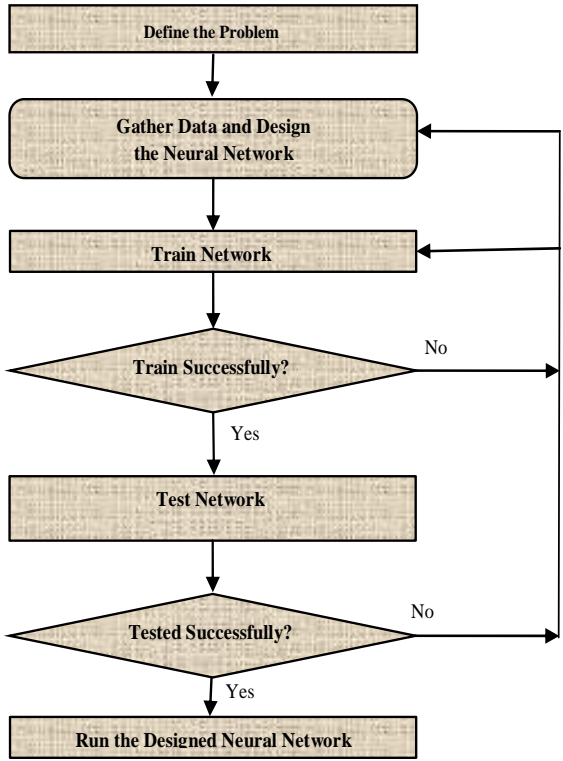

Fig. 3. Neural Network Design. **[8]**

### 6.1. Design of the Neural Network Model

Through this step, the following sequences were followed:

   **i.** Neural Network Simulation Software Selection

Many design software are used for creating neural network models. As stated earlier in the previous studies phase, many researchers used Neural Network Software in construction management in general. In this research, the N-Connection Professional Software Version 2.0 was used to develop the Neural Network Model.

This application software is very easy to use and its predicting accuracy is very high compared to other software program. It is compatible with Microsoft Windows. The N-Connection uses the back propagation algorithm in its engine. The past researches proved that the back-propagation rule is a suitable learning rule for most problems. It is the most commonly used technique for solving estimation and prediction problems **[16]**.

Firstly, in order to design the neural network model the (N-Connection V2.0) guidelines will be used for assistance. Moreover, to verify this research work the trial and error process was used to obtain the best structure of the model. During this procedure if the network is not trained satisfactory, adding or removing of hidden layers and hidden nodes will be performed until an acceptable model structure is reached, that can predict the percentage of site overhead cost with an acceptable error limit. The learning rate, training and testing tolerance are fixed by the N-Connection V 2.0 automatically **[16]**.

   **ii.** Determining the Best Network Architecture

There are two questions in neural network designing that have no precise answers because they are application-dependent: How much data do you need to train a network**? And,** how many hidden layers and nodes are the best numbers to use**?** In general, the more facts and the fewer hidden layers and hidden nodes that you can use, is the better **[16]**. There is a subtle relationship between the number of facts and the number of hidden layers/nodes. Having too few facts or too many hidden layers/nodes can cause the network to "Memorize". When this happens, it performs well during training but tests poorly **[16]**. The network architecture refers to the number of hidden layers and the number of nodes within each hidden layer **[16]**. The two guidelines that are discussed in the following section can be used in answering the last two questions **[8]**.

   **iii.** Determining the Number of Hidden Layers/Nodes

Hidden layer is a layer of neurons in an artificial neural network that does not connect to the outside world but connects to other layers of neurons **[16]**.

Hegazy et al. (1995), stated that one hidden layer with a number of hidden neurons as one-half of the total input and output neurons is suitable for most applications, but due to the ease of changing the network architecture during training, an attempt will be performed to verify this research work, through finding the network structure that generates the minimum RMS value for the given problem output parameters **[9]**.

Before starting to build, train and validate the network model, there are two parameters that should be well defined to have a good training manner. These parameters are:

   **1.** Training and Testing Tolerance

Training and testing tolerance is a value that specifies how accurate the neural network's output must be considered correct during training and testing. The most meaningful tolerance is specified as a percentage of the output range, rather than the output value **[16]**.

A tolerance of 0.1 means that the output value must be within 10% of the range of the output to be considered correct. Selecting a tolerance that is too loose (large) or too tight (small) can have an impact on the network's ability to make predictions. It is important that the selected tolerance will give responses close enough to the pattern to be useful. However, it is not always possible for Neural Connection V2.0 to train if it begins with a very small tolerance. In this study the tolerance is set by the program to (0.1).

2. Learning Rate

The learning rate specifies how large an adjustment Neural Connection will make to the connection strengths when it gets a fact wrong. Reducing the learning rate may make it possible to train the network to a smaller tolerance. The learning rate pattern is automatically set by the Neural Connection 2.0 Software program in a way that maximizes the performance of the program to achieve the best results.

**iv.** Training the Network

Training the network is a process that uses one of several learning methods to modify weight, or connection strengths. All trial models experimented in this study was trained in a supervised mode by a back-propagation learning algorithm. A training data set is presented to the network as inputs, and the outputs are calculated. The differences between the calculated outputs and the actual target output are then evaluated and used to adjust the network's weights in order to reduce the differences. As the training proceeds, the network's weights are continuously adjusted until the error in the calculated outputs converges to an acceptable level. The back-propagation algorithm involves the gradual reduction of the error between model output and the target output. Hence, it develops the input to output mapping by minimizing a root mean square error (RMS) that is expressed in the equation (**1**) **[16]**:

- Equation (**1**):

$$RMS = \sqrt{\sum_{i=1}^{n}(O_i - P_i)^2 / n}$$

Where **n** is the number of samples to be evaluated in the training phase, $O_i$ is the actual output related to the sample i (i=1...**n**), and $P_i$ is the predicted output. The training process should be stopped when the mean error remains unchanged. The training file has (90%) of the collected facts, i.e. has **47** facts (Projects). These facts are used to train and validate the network **[11]**.

**v.** Testing the Network

Testing the network is essentially the same as training it, except that the network is shown facts it has never seen before, and no corrections are made. When the network is wrong, it is important to evaluate the performance of the network after the training process. If the results are good, the network will be ready to use. If not, this means that it needs more or better data or even re-designs the network. A part of the collected facts (data) around (10%), i.e. 5 facts (projects) is set aside randomly from the set of training facts (projects) **[11]**. Then these facts are used to test the ability of the network to predict a new output where the absolute difference is calculated for each test project outcome by the equation (**2**) **[16]**:

- Equation (**2**):

$$\text{Absolute Difference} = \left(\frac{\text{Real Life Target Outcome} - \text{Predicted Model Outcome}}{\text{Real Life Target Outcome}} \times 100\right)$$

An absolute difference of 10 means that there is a 10 percent difference between the models predicted outcome value and the actual real life outcome value for that given project. This difference can be positive or negative difference (i.e. absolute difference range = ±10) and that must be clearly stated when testing phase is completed for it represents one of the main features of the constructed Neural Network Model characteristics **[16]**.

**vi.** Creating Data File for Neural Connection

N-Connection 2.0 is a tool that allows creating definition, training fact, and testing facts. The database that feeds into the Excel file consists of 47 examples of building construction site overhead costs percentage for projects constructed during the period 2002 to 2009 in Egypt, and 5 examples will be set aside for the final best model testing. The Neural Connection 2.0 program will need around 34 (73%) of the facts for training, which are the calculated minimum needed number of facts for the program to train properly, which leaves 13 of the facts for validation **[11]**.

**vii.** Determining the Best Structure for the Model

The characteristics of the model learning rule, training and testing tolerance is set automatically by the program. The variables that the program requires setting during the design stage are **[16]**:

1. Number of Hidden Layers (the program accepts up to two Hidden Layers);
2. Number of Hidden Nodes in each Layer; and
3. Type of Transfer Function in each layer.

The program is generated through the following sequence of alterations and selecting the model structure that provides the minimum RMS value [11]:
1. One Hidden Layer with Sigmoid Transfer Function; (Table 2A)
2. One Hidden Layer with Tangent Transfer Function; (Table 2B)
3. Two Hidden Layers with Sigmoid Transfer Function in each; (Table 2C)
4. Two Hidden Layers with Tangent Transfer Function in each; (Table 2D)

Table 2A: Experiments for Determining the Best Model

| Model No. | Input Nodes | Output Node | No. of Hidden Layers | No. of Hidden Nodes In 1st Layer | No. of Hidden Nodes In 2nd Layer | Absolute Difference % | RMS |
|---|---|---|---|---|---|---|---|
| 1 | 10 | 1 | 1 | 3 | 0 | 7.589891 | 0.900969 |
| 2 | 10 | 1 | 1 | 4 | 0 | 5.491507 | 0.602400 |
| 3 | 10 | 1 | 1 | 5 | 0 | 8.939657 | 1.046902 |
| 4 | 10 | 1 | 1 | 6 | 0 | 7.766429 | 0.932707 |
| 5 | 10 | 1 | 1 | 7 | 0 | 4.979286 | 0.535812 |
| 6 | 10 | 1 | 1 | 8 | 0 | 5.818345 | 0.647476 |
| 7 | 10 | 1 | 1 | 9 | 0 | 4.947838 | 0.579932 |
| 8 | 10 | 1 | 1 | 10 | 0 | 8.887463 | 1.039825 |
| 9 | 10 | 1 | 1 | 11 | 0 | 4.858645 | 0.507183 |
| 10 | 10 | 1 | 1 | 12 | 0 | 5.352388 | 0.651948 |
| **11** | **10** | **1** | **1** | **13** | **0** | **2.476118** | **0.276479** |
| 12 | 10 | 1 | 1 | 14 | 0 | 2.857856 | 0.428663 |
| 13 | 10 | 1 | 1 | 15 | 0 | 4.074554 | 0.478028 |
| 14 | 10 | 1 | 1 | 20 | 0 | 8.065637 | 1.050137 |

i.e. Model trials from 1 to 14 has a Sigmoid transfer function.

The first fourteen model trails illustrated that the RMS and Absolute Difference values changed as the number of hidden nodes in the single hidden layer increased in a nonlinear relationship, where the lowest RMS value of 0.276479 and a corresponding Absolute Difference value of 2.476118 were achieved in the eleventh trial where there were thirteen hidden nodes in the single hidden layer with a sigmoid transfer function. On the other side highest RMS value of 1.050137 and the corresponding Absolute Difference value of 8.065637 were achieved in the fourteenth trial when there was twenty hidden nodes in the single hidden layer with a sigmoid transfer function. For the remaining twelve model trails the RMS and Absolute Difference values changed consecutively within the above mentioned ranges for each model trial.

Table 2B: Experiments for Determining the Best Model

| Model No. | Input Nodes | Output Node | No. of Hidden Layers | No. of Hidden Nodes In 1st Layer | No. of Hidden Nodes In 2nd Layer | Absolute Difference % | RMS |
|---|---|---|---|---|---|---|---|
| 15 | 10 | 1 | 1 | 3 | 0 | 3.809793 | 0.490956 |
| 16 | 10 | 1 | 1 | 4 | 0 | 5.666974 | 0.703804 |
| 17 | 10 | 1 | 1 | 5 | 0 | 3.813867 | 0.425128 |
| 18 | 10 | 1 | 1 | 6 | 0 | 5.709665 | 0.709344 |
| 19 | 10 | 1 | 1 | 7 | 0 | 5.792984 | 0.634338 |
| 20 | 10 | 1 | 1 | 8 | 0 | 2.952316 | 0.343715 |
| 21 | 10 | 1 | 1 | 9 | 0 | 5.629162 | 0.655106 |
| 22 | 10 | 1 | 1 | 10 | 0 | 3.544173 | 0.387283 |
| 23 | 10 | 1 | 1 | 11 | 0 | 5.578666 | 0.686378 |
| 24 | 10 | 1 | 1 | 12 | 0 | 5.772656 | 0.701365 |
| 25 | 10 | 1 | 1 | 13 | 0 | 3.582526 | 0.380564 |
| 26 | 10 | 1 | 1 | 14 | 0 | 4.614612 | 0.515275 |
| 27 | 10 | 1 | 1 | 15 | 0 | 4.806596 | 0.641098 |
| 28 | 10 | 1 | 1 | 20 | 0 | 7.005237 | 0.826699 |

i.e. Model trials from 15 to 28 has a Tangent transfer function.

The model trails from 15 to 28 where there is one hidden layer, illustrated that the RMS and Absolute Difference values changed as the number of hidden nodes/hidden layer changed in a nonlinear relationship, where the lowest RMS value of 0.343715 and a corresponding Absolute Difference value of 2.952316 were achieved in the twentieth model trial when there was eight (8) hidden nodes in the single hidden layer. On the other side, with a tangent transfer function, the highest RMS value of 0.826699 and the corresponding Absolute Difference value of 7.005237 were achieved in the twenty eighth model trial when there were twenty hidden nodes in the single hidden layer. The remaining values changed consecutively within the above mentioned ranges for each model trial.

Table 2C: Experiments for Determining the Best Model

| Model No. | Input Nodes | Output Node | No. of Hidden Layers | No. of Hidden Nodes In 1st Layer | No. of Hidden Nodes In 2nd Layer | Absolute Difference % | RMS |
|---|---|---|---|---|---|---|---|
| 29 | 10 | 1 | 2 | 2 | 1 | 9.919941 | 1.519966 |
| 30 | 10 | 1 | 2 | 2 | 2 | 5.170748 | 0.581215 |
| 31 | 10 | 1 | 2 | 3 | 1 | 10.374248 | 1.413138 |
| 32 | 10 | 1 | 2 | 3 | 2 | 11.167767 | 1.687072 |
| 33 | 10 | 1 | 2 | 3 | 3 | 8.013460 | 1.140512 |
| 34 | 10 | 1 | 2 | 4 | 1 | 5.679721 | 0.643957 |
| 35 | 10 | 1 | 2 | 4 | 2 | 5.577789 | 0.617385 |
| 36 | 10 | 1 | 2 | 4 | 3 | 5.448696 | 0.598400 |
| 37 | 10 | 1 | 2 | 4 | 4 | 4.079718 | 0.492011 |
| 38 | 10 | 1 | 2 | 5 | 3 | 4.191063 | 0.574500 |
| 39 | 10 | 1 | 2 | 5 | 4 | 6.024062 | 0.723419 |
| 40 | 10 | 1 | 2 | 5 | 5 | 5.322466 | 0.654373 |
| 41 | 10 | 1 | 2 | 6 | 4 | 7.257790 | 0.804202 |
| 42 | 10 | 1 | 2 | 6 | 5 | 5.158298 | 0.567479 |
| 43 | 10 | 1 | 2 | 6 | 6 | 5.270355 | 0.545017 |

i.e. Model trials from 29 to 43 has a Sigmoid transfer function for both hidden layers.

The model trails from 29 to 43 illustrated that the RMS and Absolute Difference values changed as the number of hidden nodes per each hidden layer increased in a nonlinear relationship, where the lowest RMS value of 0.492011 and a corresponding Absolute Difference value of 4.079718 were achieved in the model trial number (37) when there were two hidden layers with four hidden nodes in each layer and having a sigmoid transfer function. Contrarily, the highest RMS value of 1.687072 and the corresponding Absolute Difference value of 11.167767 were achieved in the model trial number (32) when there were two hidden layers with three hidden nodes in the fist layer and two hidden nodes in the second hidden layer and having a sigmoid transfer function. For the remaining thirteen model trails the RMS and Absolute Difference values changed consecutively within the above mentioned ranges for each model trial having a sigmoid function in each layer.

Table 2D: Experiments for Determining the Best Model

| Model No. | Input Nodes | Output Node | No. of Hidden Layers | No. of Hidden Nodes In 1st Layer | No. of Hidden Nodes In 2nd Layer | Absolute Difference % | RMS |
|---|---|---|---|---|---|---|---|
| 44 | 10 | 1 | 2 | 2 | 1 | 4.364562 | 0.499933 |
| 45 | 10 | 1 | 2 | 2 | 2 | 3.551318 | 0.380629 |
| 46 | 10 | 1 | 2 | 3 | 1 | 4.787220 | 0.493240 |
| 47 | 10 | 1 | 2 | 3 | 2 | 6.267891 | 0.852399 |
| 48 | 10 | 1 | 2 | 3 | 3 | 6.515138 | 0.829739 |
| 49 | 10 | 1 | 2 | 4 | 1 | 3.458081 | 0.481580 |
| 50 | 10 | 1 | 2 | 4 | 2 | 9.249286 | 1.158613 |
| 51 | 10 | 1 | 2 | 4 | 3 | 4.735680 | 0.552350 |
| 52 | 10 | 1 | 2 | 4 | 4 | 7.445228 | 0.991062 |
| 53 | 10 | 1 | 2 | 5 | 3 | 7.729862 | 1.105441 |
| 54 | 10 | 1 | 2 | 5 | 4 | 9.807989 | 1.180131 |
| 55 | 10 | 1 | 2 | 5 | 5 | 6.060798 | 0.657344 |
| 56 | 10 | 1 | 2 | 6 | 4 | 3.213154 | 0.355932 |
| 57 | 10 | 1 | 2 | 6 | 5 | 4.381631 | 0.490479 |
| 58 | 10 | 1 | 2 | 6 | 6 | 4.731568 | 0.502131 |

i.e. Model trials from 44 to 58 has a Tangent transfer function for both hidden layers.

The model trails from 44 to 58 illustrated that the RMS and Absolute Difference values changed as the number of hidden nodes per each hidden layer increased in a nonlinear relationship, where the lowest RMS value of 0.355932 and a corresponding Absolute Difference value of 3.213154 were achieved in the model trial number (56), when there was two hidden layers with six hidden nodes in the first hidden layer and four hidden nodes in the second hidden layer and with a tangent transfer function in each layer. On the other side, the highest RMS value of 1.180131 and the corresponding Absolute Difference value of 9.807989 were achieved in the model trial number (54) when there was two hidden layers with five hidden nodes in the fist layer and four hidden nodes in the second hidden layer and with a tangent transfer function in each layer. For the remaining thirteen model trails the RMS and Absolute Difference values changed consecutively within the above mentioned ranges for each and with a sigmoid function in each layer **[11]**.

The recommend model for this prediction problem is that with the least RMS value from all the fifty-eight trails and error process **[16]**. This is trial number eleven **[11]**.

As a result, from training phase the characteristics of the satisfactory Neural Network Model that was obtained through the trail and error process are presented in (Table 3) and (Fig. 4).

Model Trial Number Eleven with the following Eight Design Parameters, which are **[11]**:

1. Input layer with **10** Neurons (nodes);
2. One hidden layer with **13** Neurons (nodes);
3. Output layer with **1** Neuron (node);
4. With a Sigmoid Transfer Function;
5. Learning rate automatically adjusted by the program;
6. Training Tolerance = **0.10** (Adjusted by Program);
7. Root Mean Square Error = **0.276479**;
8. Absolute Mean Difference % = **2.476118**.

Table 3: Characteristics of the Best Model

| Model | No. of input nodes | No. of hidden layers | No. of nodes/ hidden layer | LR | TF | No. of output nodes | RMS |
|---|---|---|---|---|---|---|---|
| **11** | **10** | **1** | **13** | **Back propagation** | **Sigmoid function** | **1** | **0.276479** |

LR: Learning Rule; TF: Transfer Function; RMS: Root Mean Square Error.

**viii.** Testing the Validity of the Model

To evaluate the predictive performance of the network, the five projects that were previously randomly selected and reserved for testing from the total collected projects are introduced to the best model without the percentage of their site overhead cost for testing the prediction ability of the designed ANN-program.

The model will predict the percentage of building construction projects site overhead costs for projects constructed in Egypt. The predicted percentage will be compared to the real life projects percentage (stored outside the program). The difference between them will be calculated if it is equal or under the value of the designed model's Absolute Difference, then it is considered to be a correct prediction attempt. If it exceeds the value of the designed model Absolute Difference, then it is considered to be a wrong prediction attempt, (Table 4) presents the actual and predicted percentages for the test sample.

The model correctly predicted four from the five testing projects sample which is 80% of the test sample. The wrongly predicted project had a positive difference between the value of predicted percentage from the model output and the real life percentage for the same project equal to (+) 4.620294427%. This means that the predicted outcome is greater than the actual real life project value by this percentage value **[11]**. Such percentage is found to be acceptable; program user's manual, because the difference between the predicted program outcome for this project and the real life outcome for the same project is less than five percent (5%) which is found by the program to be very small (under 10%) and acceptable. The program (user's manual) clearly dictates to regard small differences and accept any sample difference that small to be a correct sample **[16]**. But even if the model's correct predicted outcome is taken to be (80%) that will still be considered as a very high and the model is accepted **[8]**.

Table 4: Actual and Predicted Percentage of Building Site Overhead for the Test Sample.

| Project No. | Actual real life percentage | Network output (predicted percentage) | Absolute difference % | Comments |
|---|---|---|---|---|
| **1** | 8.13 | 8.32294 | ( - ) 2.373185732 | **Correct** |
| **2** | 9.51 | 9.07061 | ( + ) 4.620294427 | **Wrong** |
| **3** | 10.86 | 10.59704 | ( + ) 2.421362799 | **Correct** |
| **4** | 10.84 | 11.11394 | ( - ) 2.427121771 | **Correct** |
| **5** | 11.43 | 11.3421 | ( + ) 0.769028871 | **Correct** |

As it is clear the correct predicted model outputs of the percentage of site overhead costs differ from the actual real life project percentage of site overhead costs value with a value under (±2.476%) which is the designed model absolute difference%, which is assumed to be acceptable.

This demonstrates a very high accuracy for the proposed model and the viability of the neural network as a powerful tool for modeling the assessment of the building construction site overhead cost percentage for projects constructed in Egypt **[11]**.

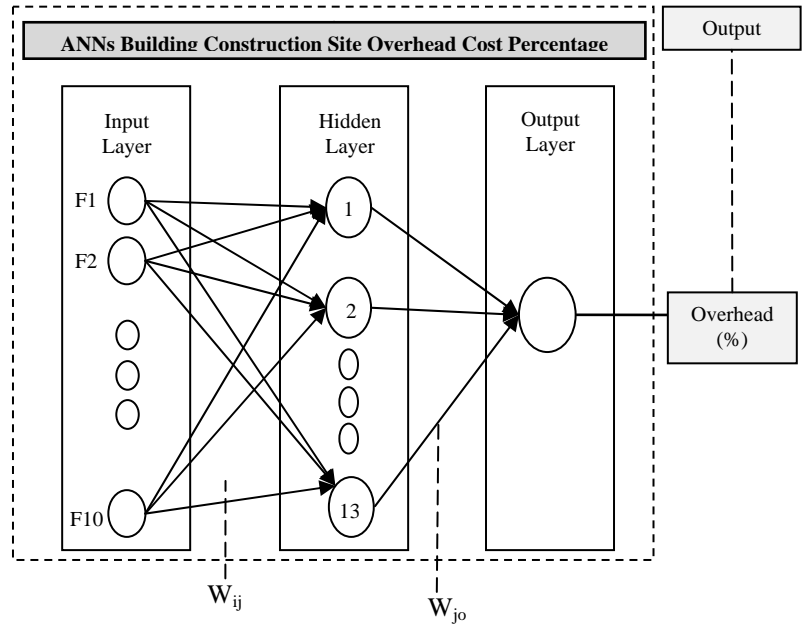

Fig. 4. Structure of the Best Model. **[11]**

## 7. Summary

Construction firms should carefully examine contract conditions and perform all the necessary precautions to make sure that project site overhead costs factors are properly anticipated for and covered within the total tender price. The study conducted a survey that investigated the factors affecting project's site overhead cost for building construction projects in the first and second categories of construction companies. An ANN model was developed to predict the percentage of site overhead cost for building construction projects in Egypt during the tendering process. A sample of building projects was selected as a test sample for this study. The impacts of different factors on the site overhead costs were deeply investigated. The survey results illustrated that site overhead costs are greatly affected by many factors. Among these factors come project type, size, location, site conditions and the construction technology. All of these factors make the detailed estimation of such overhead costs a more difficult task.

Hence, it is expected that a lump-sum assessment for such cost items will be a more convenience, easy, highly accurate, and quick approach. Such approach should take into consideration the different factors that affect site overhead cost. It was found that an ANN-Based Model would be a suitable tool for site overhead cost assessment.

## 8. CONCLUSIONS

The following conclusions are drawn from this research:
1. Through literature review potential factors that influence the percentage of site overhead costs for building construction projects were identified. Ten factors were identified;
2. The analysis of the collected data gathered from fifty-two real-life building construction projects from Egypt illustrated that project's duration, total contract value, projects type, special site preparation needs and project's location are identified as the top five factors that affect the value of the percentage of site overhead costs for building construction projects in Egypt;
3. Nature of the client, type of the contract and contractor-joint venture are the lowest affecting factors in the percentage of site overhead costs for building construction projects in Egypt;
4. A satisfactory Neural Network model was developed through fifty-eight experiments for predicting the percentage of site overhead costs for building construction projects in Egypt for the future projects. This model consists of one input layer with ten neurons (nodes), one hidden layer having thirteen hidden nodes with a sigmoid transfer function and one output layer. The learning rate of this model is set automatically by the N-Connection V2.0 while the training and testing tolerance are set to 0.1;

5. The results of testing for the best model indicated a testing root mean square error (RMS) value of 0.276479; and
6. Testing was carried out on five new facts (Projects) that were still unseen by the network. The results of the testing indicated an accuracy of (**80**%). As the model wrongly predicted the percentage of site overhead costs for only one project (**20**%) from the testing sample.

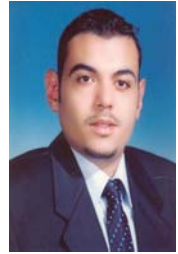

**Ismaail Yehia Aly ElSawy**, has received his M.Sc. and B.Sc. degrees in Construction and Building Engineering, College of Engineering and Technology, from Arab Academy for Science, Technology and Maritime Transport, Alexandria, Egypt, 2010 and 2002. He Joined in December (2004) the Egyptian Ministry of Electricity and Power as a Research Engineer in the Ministries National Research Center. He then joined the academic field in September (2008), as a Demonstrator (B.Sc.) then Assistant lecturer (M.Sc.) at the Civil Engineering Department, Thebes Higher Institute of Engineering. He has published more than 6 research papers in International/National Journals and Refereed International Conferences. He is interested in the implementation of Artificial Intelligence in Construction Project Management, and Construction Projects Financial Management.